\documentclass[10pt,twocolumn,letterpaper]{article}

\usepackage{iccv}              
\usepackage{float}
\definecolor{iccvblue}{rgb}{0.21,0.49,0.74}
\usepackage[pagebackref,breaklinks,colorlinks,allcolors=iccvblue]{hyperref}

\def\paperID{*****} 
\def\confName{ICCV}
\def\confYear{2025}

\title{%
\centering
{\bfseries\Large DriveExplorer: Images-Only Decoupled 4D Reconstruction with Progressive Restoration for Driving View Extrapolation}%
}

\author{
\begin{tabular}{c}
\textbf{Yuang Jia} \quad
\textbf{Jinlong Wang} \quad
\textbf{Jiayi Zhao} \quad
\textbf{Chunlam Li} \quad
\textbf{Shunzhou Wang} \quad
\textbf{Wei Gao\thanks{Corresponding Author: Wei Gao}} \\
Guangdong Provincial Key Laboratory of Ultra High Definition Immersive Media Technology, \\
School of Electronic and Computer Engineering, Shenzhen Graduate School, Peking University \\
yuangjia8@gmail.com \quad gaowei262@pku.edu.cn \\
\end{tabular}
}

\begin{document}
\vspace{-40pt} 
\maketitle
\vspace{-30pt}

\begin{abstract}
This paper presents an effective solution for view extrapolation in autonomous driving scenarios. Recent approaches focus on generating shifted novel view images from given viewpoints using diffusion models. However, these methods heavily rely on priors such as LiDAR point clouds, 3D bounding boxes, and lane annotations, which demand expensive sensors or labor-intensive labeling, limiting applicability in real-world deployment. In this work, with only images and optional camera poses, we first estimate a global static point cloud and per-frame dynamic point clouds, fusing them into a unified representation. We then employ a deformable 4D Gaussian framework to reconstruct the scene. The initially trained 4D Gaussian model renders degraded and pseudo-images to train a video diffusion model. Subsequently, progressively shifted Gaussian renderings are iteratively refined by the diffusion model, and the enhanced results are incorporated back as training data for 4DGS. This process continues until extrapolation reaches the target viewpoints. Compared with baselines, our method produces higher-quality images at novel extrapolated viewpoints.
\end{abstract}
\vspace{-10pt}    
\vspace{-10pt}
\section{Introduction}
\label{sec:intro}
\vspace{-5pt}
Recently, closed-loop simulation has emerged as one of the central focuses in autonomous driving research. Existing approaches to closed-loop simulation can be broadly categorized into two paradigms: The first relies on explicit reconstruction methods, such as Neural Radiance Fields (NeRF) ~\cite{streetsurf,yang2023unisim} and 3D Gaussian Splatting (3DGS) ~\cite{chen2024omnire,yan2024street}. Although these methods are capable of producing high-quality renderings near the original viewpoints, their performance deteriorates in complex or dynamic scenes due to limitations in the density and uniformity of the initial data sampling. The second paradigm ~\cite{drivingsphere} leverages diffusion models to implicitly generate images from novel viewpoints. These approaches tend to produce visually sharp results and preserve most spatio-temporal consistency; however, they often suffer from varying degrees of distortion. Moreover, neither category of methods is able to achieve high-quality generation under large viewpoint shifts: explicit reconstruction produces severe artifacts and misalignments, while a single diffusion-based generation approach introduces substantial distortions.
\begin{figure}[t]
    \centering
    \vspace{-15pt}
    \includegraphics[width=\columnwidth]{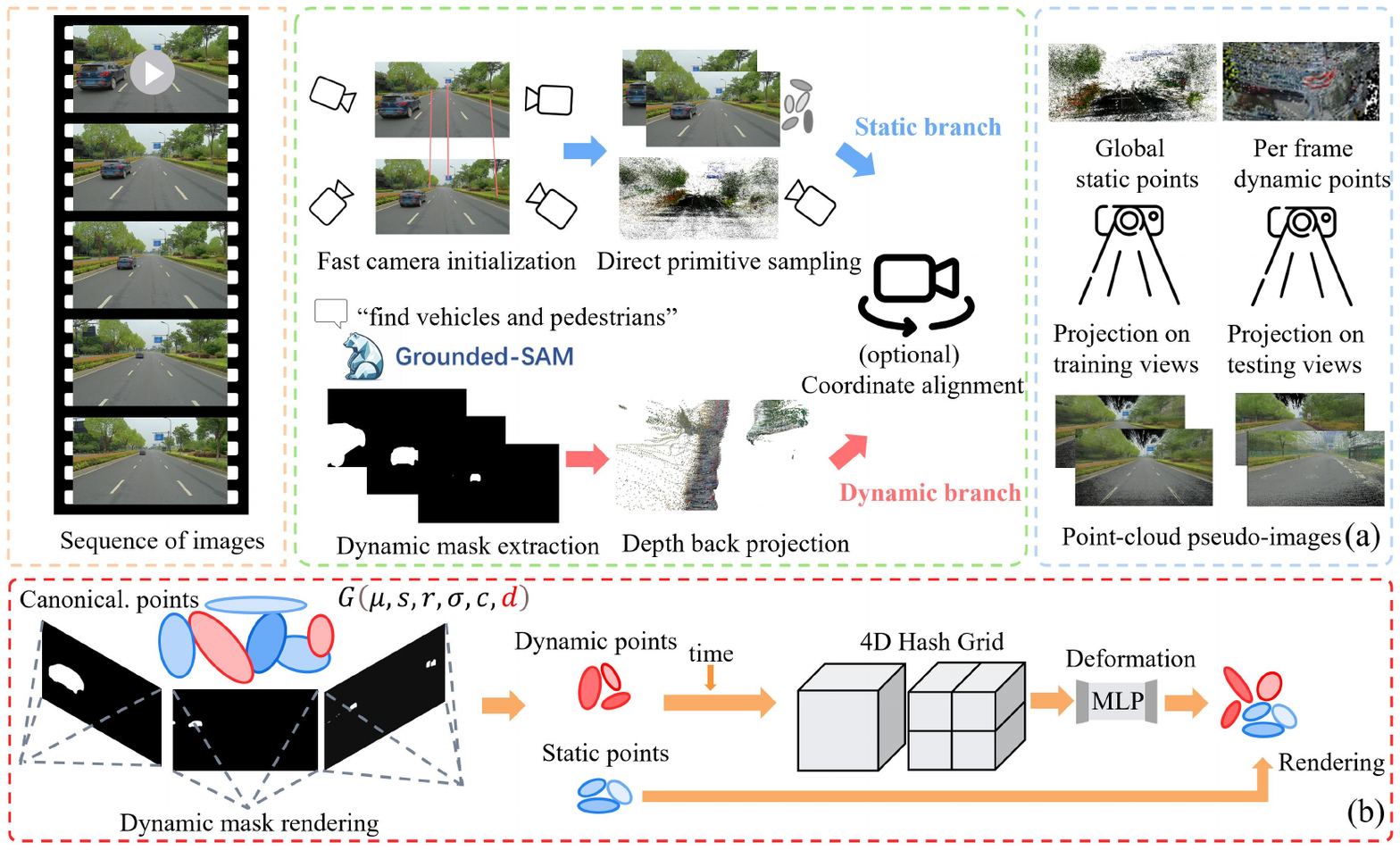} 
    \captionsetup{justification=raggedright, singlelinecheck=false} 
    \vspace{-30pt}
    \caption{(a) Data processing pipeline. We decouple static and dynamic point clouds directly from images, and project the fused global point cloud onto both the original viewpoints and shifted test viewpoints. (b) Gaussian training pipeline. We first train the dynamic Gaussians to render masks at the original viewpoints, ensuring the ability to generate dynamic masks from any viewpoint thereafter. We then deform the dynamic Gaussians using a 4D grid and couple them with the static Gaussians.} 
    \vspace{-10pt}
    \label{data-process} 
\end{figure}

To introduce high-quality shifted viewpoints into closed-loop simulation, thus enlarging the perceptual coverage of the simulated environment, recent work has proposed specialized training strategies for handling viewpoint shifts. For example, several methods \cite{guo2025dist,zou2025mudg} enhance the generative capacity at novel points of view by augmenting the input conditions of video diffusion models (VDMs). DriveDreamer4D \cite{zhao2025drivedreamer4d} instead employs a world model as a generator to directly synthesize shifted view images to train Gaussian-based representations. Building upon this, ReconDreamer \cite{ni2025recondreamer} introduces progressive training to support novel-view rendering under multi-lane shifts. In addition, ReconDreamer++ \cite{zhao2025recondreamer++} refines this direction by decoupling the scene into multiple components for rendering. However, these methods heavily rely on annotated data such as 3D bounding boxes, lane markings, and LiDAR point clouds, which significantly constrains their practicality for real-world deployment.

To address the challenge of large-scale novel view synthesis using only images (with optional camera poses), we propose a method that separately estimates dense static point clouds and per-frame dynamic point clouds and employs 4D Gaussian Splatting to perform dynamic modeling of the estimated scene. The general data processing and Gaussian training pipeline are illustrated in Fig. \ref{data-process}. Specifically, we render under-trained Gaussians into degraded video sequences, and use the corresponding dense point clouds together with dynamic masks as conditions to train a video diffusion model. During inference, progressively shifted test viewpoints are generated through the trained diffusion model, and the results are iteratively incorporated as new ground truth for retraining 3DGS, until the extrapolation reaches the target viewpoints. Our approach introduces a new images-only training paradigm, and achieves significant improvements over various baselines in terms of PSNR, SSIM, and PIPS metrics.

\vspace{-10pt}
\section{Related Work}
\label{sec:formatting}
\vspace{-5pt}
\noindent\textbf{Driving Scene Reconstruction.}
Recent methods for autonomous driving have advanced rapidly, with many adopting NeRF-based pipelines. URF \cite{rematas2022urban} combines LiDAR and RGB images, Block-NeRF \cite{tancik2022block} partitions large scenes for efficient training, and Neural Scene Graphs \cite{ost2021neural} structure dynamic environments. Self-supervised approaches \cite{yang2023emernerf} further improve handling of dynamics. In parallel, 3DGS-based works such as StreetGaussian \cite{yan2024street}, DrivingGaussian \cite{zhou2024drivinggaussian}, and Omnire \cite{chen2024omnire} target static/dynamic modeling with specialized designs. However, these methods still rely on auxiliary priors (e.g., 3D boxes, lane annotations, LiDAR), limiting deployment. By contrast, our method learns solely from images and enables high-quality lane-shifted novel view synthesis in dynamic driving scenes.

\noindent\textbf{Lane-Shifted Novel View Synthesis.} With advances in diffusion models and world models, an increasing number of studies have shifted their focus from rendering within the original driving viewpoints to synthesizing images under lane-shifted perspectives. Specifically, DriveDreamer4D \cite{zhao2025drivedreamer4d} is the first to employ a world model \cite{wang2024drivedreamer} to directly produce new trajectory videos, enabling joint training of reconstruction modules. In contrast, ReconDreamer \cite{ni2025recondreamer} utilizes video synthesis techniques to enhance rendering quality under unseen viewpoints. Extending these ideas, ReconDreamer++ \cite{zhao2025recondreamer++} introduces a trainable spatial deformation strategy that aligns generated views with sensor-captured observations, effectively narrowing the domain discrepancy. Nevertheless, these approaches essentially extend multi-source Gaussian reconstruction, and thus still face limitations when applied to real-world scenarios. In contrast, our method leverages learned Gaussian-based dynamic masks together with point-cloud pseudo-images as conditions for the diffusion model, without requiring any additional priors.
\vspace{-20pt}
\section{Our Algorithm}
\vspace{-5pt}
\begin{figure}[t]
    \centering
    \vspace{-40pt}
    \includegraphics[width=\columnwidth]{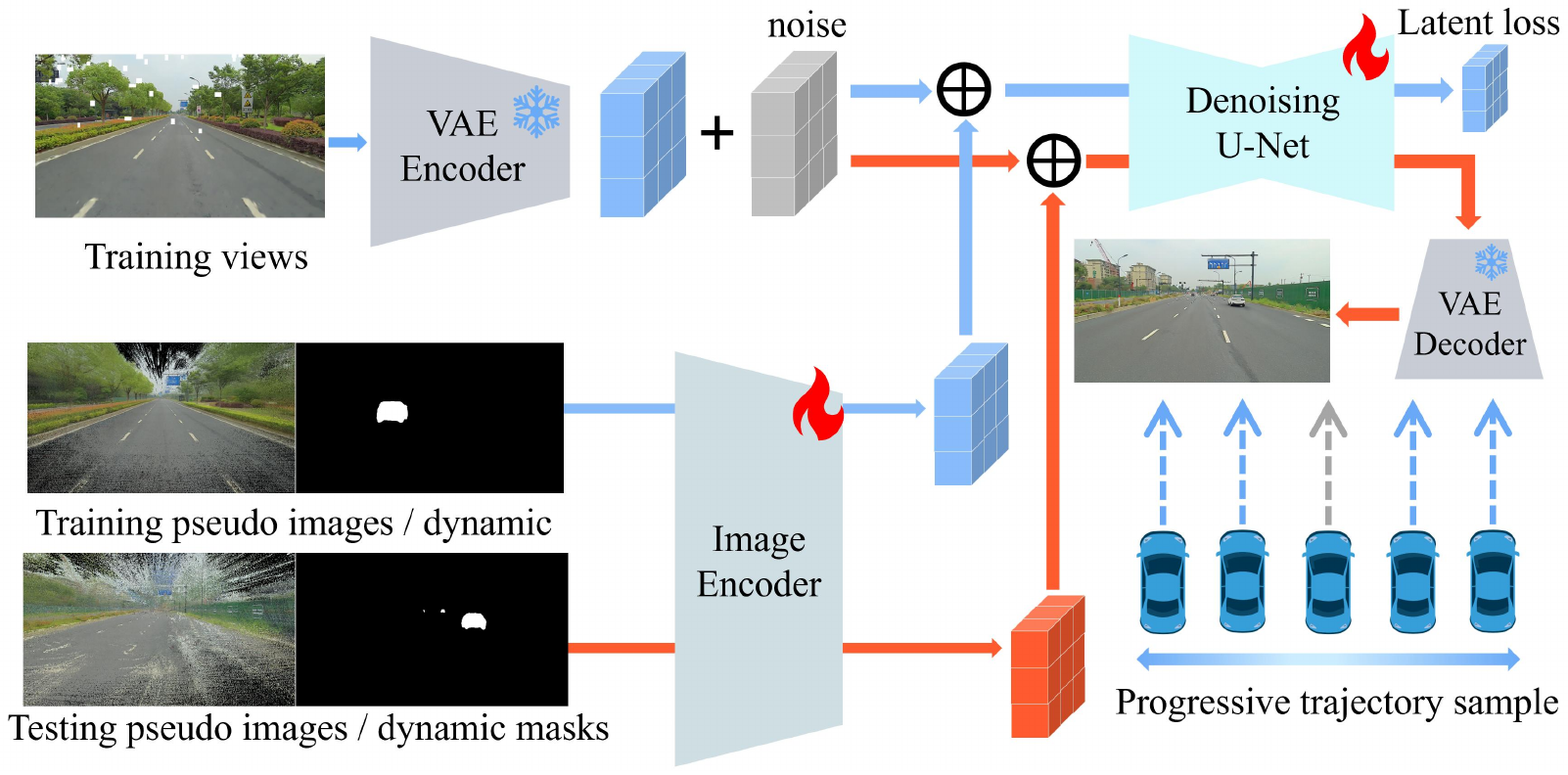} 
    \captionsetup{justification=raggedright, singlelinecheck=false} 
    \vspace{-40pt}
    \caption{Training pipeline of the video diffusion model. The pseudo-images and Gaussian dynamic mask obtained from Fig. \ref{data-process} are used as conditional inputs, concatenated with the latent features, and then passed to the denoising U-Net.} 
    \vspace{-15pt}
    \label{pipeline} 
\end{figure}

\noindent\textbf{Mask-Aware Gaussian Training.} We initialize the Gaussians using the point clouds estimated in the previous subsection. For each dynamic Gaussian, we assign an additional parameter \(d\). For every pixel, we apply \(\alpha\)-composition with the parameter \(d\) through a sigmoid function to render a dynamic value \(\hat{M}(x)\) at location \(x\), as illustrated in Eq. \ref{eq1}. 
\begin{equation}
    \hat{M}(x) = \operatorname{Sigmoid}\big( 
    \sum_{i=1} d_{i}\alpha_{i}(x) 
    \prod_{j=1}^{i-1} (1-\alpha_{j}(x))\big),
    \tag{1}
    \label{eq1}
\end{equation}
where $\alpha_i$ and $\alpha_j$ denote the opacity of the $i$-th and $j$-th Gaussian respectively. Therefore, the estimation of the dynamic property for each Gaussian point is achieved by formulating it as a binary classification task and minimizing the corresponding cross-entropy objective:
\begin{equation}
    \mathcal{L}_{d}=\mathbb{E}_{x}[\operatorname{BCE}(M(x), \hat{M}(x))],
    \tag{2}
    \label{eq2}
\end{equation}
where $M(x)$ denotes the reference pixel obtained from \cite{ren2024grounded} and BCE denotes the binary cross-entropy loss. From the above equation, the parameter \(d\) of dynamic Gaussians is continuously optimized. After optimizing the parameter \(d\), dynamic binary masks can be produced from arbitrary viewpoints. For a 4D dynamic Gaussian point \((\mu, t) \in \mathbb{R}^{4}\), we encode it with a multi-resolution hashing function \(\phi_{en}(\mu, t ; k)\), where the representation at resolution level \(k\) is written as \(\phi_{en}(\mu, t ; k) \in \mathbb{R}^{F}\). Each level’s feature is obtained by interpolating the vectors at the neighboring grid corners. By concatenating the features from all \(K\) levels, we obtain the final hash-based embedding vector:
\vspace{-5pt}
\begin{equation}
    f_{\phi}=\left[\phi_{en}(\mu, t ; 0), \phi_{en}(\mu, t ; 1) \ldots \phi_{en}(\mu, t ; K-1)\right],
    \tag{3}
    \label{eq3}
\end{equation}
Based on this representation, we employ a multi-head Gaussian deformation decoder to predict the temporal variations of position \(\mu\), scale \(s\), rotation \(r\), opacity \(\sigma\), and color coefficients \(c\) for each Gaussian. This yields updated parameters of the dynamic Gaussians at time \(t\), which are then combined with the static Gaussians for the final rendering.

\noindent\textbf{Video Diffusion Model Training with Multi-Conditional Inputs.} As shown in Fig. \ref{pipeline}, we render Gaussian representations with varying levels of degradation across different iterations to construct training samples, and initialize the Video Diffusion Model with weights from \cite{blattmann2023stable}. The training data are first encoded into the latent space using a VAE encoder, where Gaussian noise is applied. A trainable 2D image encoder is employed to separately encode the pseudo-images and dynamic masks; their embeddings are concatenated with the latent features and subsequently fed into a denoising U-Net, with the loss computed in the latent space. Following \cite{ni2025recondreamer}, we adopt a progressive shifting strategy: after generating images at shifted viewpoints, these synthesized images are treated as new ground truth to further train the Gaussians, until the shift reaches the final test viewpoints. For both the original and shifted viewpoints, we only employ the L1 and SSIM loss.
\vspace{-10pt}
\section{Experiment}
\vspace{-2pt}
\noindent\textbf{Datasets.} Our experiments are conducted on a subset of the EUVS dataset \cite{han2024extrapolated}. EUVS is specifically designed to provide ground-truth data from shifted viewpoints, which allows the computation of metrics such as PSNR, SSIM, and LPIPS. The subset we use covers three levels of shifting difficulty: one-lane shift, two-lane shift, and opposite-lane shift.

\noindent\textbf{Detailed settings of the experiment.} We use degraded Gaussian renderings at 6,000, 8,000, and 10,000 iterations as training samples and apply random masking to the sky and distant regions following \cite{ni2025recondreamer}. For each  viewpoint of different scenes, the image sequences are organized into continuous videos and fed into the video diffusion mode, where each frame is cropped to 1024 × 576. The model is trained for 150,000 iterations in total. After training the video diffusion model, novel shifted viewpoints are introduced once the 4D Gaussians have been trained to 25,000 iterations, and the images from these shifted viewpoints are fused with the original viewpoint images in a 1:1 ratio to form a new training set. The training is continued for an additional 20,000 iterations after each update. All experiments are conducted on two NVIDIA H20 GPUs.

\begin{figure}[H]
    \vspace{-10pt}
    \centering
    \includegraphics[width=\columnwidth]{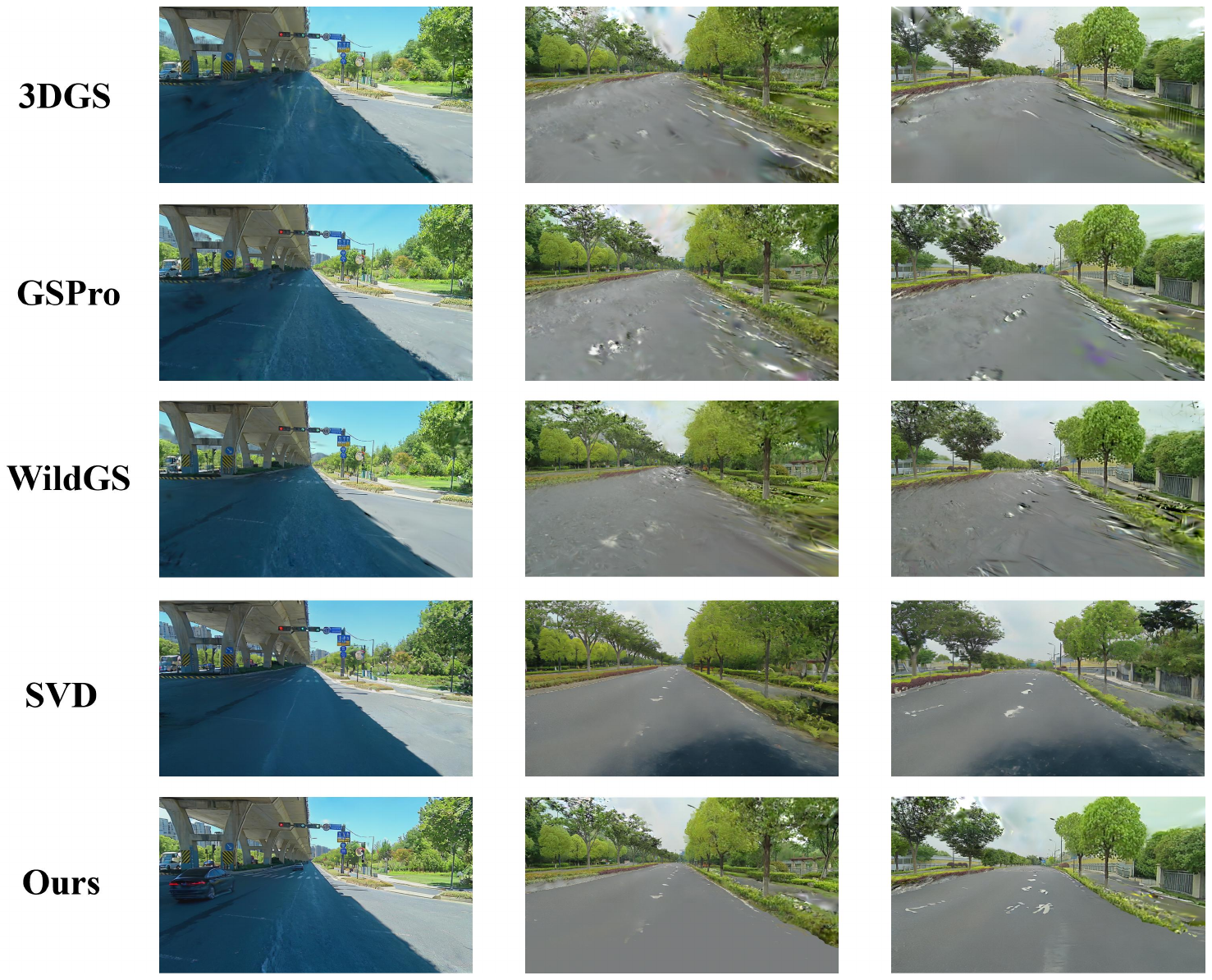}
    \captionsetup{justification=centering}
    \vspace{-20pt}
    \caption{Qualitative results on EUVS dataset.}
    \vspace{-10pt}
    \label{presentation}
\end{figure}

\noindent\textbf{Results on the EUVS Dataset.} Table \ref{results} presents a quantitative comparison between our approach and several baselines \cite{kerbl20233d,cheng2024gaussianpro,zhang2024gaussian,blattmann2023stable}. Our method outperforms existing approaches that rely solely on images as input across all evaluation metrics. Figure \ref{presentation} shows qualitative results, where it can be observed that our method produces a clearer modeling of both dynamic objects and static backgrounds.

\noindent\textbf{Ablation Study.} We further study the impacts of essential components. As seen in Table \ref{Ablation}, initializing with dense points significantly improves the overall quality of the scene reconstruction. Dynamic modeling effectively alleviates interference from moving objects, while incorporating multi-conditional inputs into the diffusion process substantially enhances the generative capability of the diffusion model.
\vspace{-10pt}
\begin{table}[h]
\caption{\textbf{Results on EUVS dataset.} The two best scores for each metric are marked in \textcolor{red}{red} and \textcolor{blue}{blue}, respectively.}
\centering
\begin{tabular}{lccc}
\toprule
\textbf{Methods} & PSNR \(\uparrow \) & SSIM \(\uparrow \) & LPIPS \(\downarrow \) \\
\midrule
3DGS \cite{kerbl20233d} & 15.691 & 0.47 & 0.564 \\
WildGS \cite{zhang2024gaussian} & 15.946 & 0.475 & 0.535 \\
GaussianPro \cite{cheng2024gaussianpro} & \textcolor{blue}{16.058} & \textcolor{blue}{0.477} & 0.520 \\
SVD \cite{blattmann2023stable} & 15.958 & 0.472 & \textcolor{blue}{0.502} \\
\textbf{Ours} & \textcolor{red}{16.519} & \textcolor{red}{0.486} & \textcolor{red}{0.442} \\
\bottomrule
\end{tabular}
\label{results}
\end{table}

\vspace{-20pt}
\begin{table}[h]
\caption{\textbf{Ablation study on EUVS dataset of our method.}}
\centering
\begin{tabular}{lccc}
\toprule
\textbf{Variant} & PSNR \(\uparrow \) & SSIM \(\uparrow \) & LPIPS \(\downarrow \) \\
\midrule
\textbf{Full Model} & \textbf{16.519} & \textbf{0.486} & \textbf{0.442} \\
w/o. dense points & 15.872 & 0.474 & 0.527 \\
w/o. dynamic deformation & 15.958 & 0.473 & 0.537 \\
w/o. multi conditions & 16.218 & 0.481 & 0.457 \\
\bottomrule
\end{tabular}
\label{Ablation}
\end{table}
\vspace{-10pt}

\vspace{-10pt}
\section{Conclusion}
\vspace{-5pt}
In this paper, we propose a novel images-only approach that achieves high-quality rendering under shifted viewpoints. The effectiveness of the proposed framework is further validated through ablation studies, which highlight the importance of its key components. Nevertheless, certain limitations remain: due to inaccuracies in the estimated dense point cloud depth, it is not feasible to incorporate depth-supervision loss, resulting in suboptimal reconstruction in some regions, particularly around dynamic objects. Addressing this issue constitutes an important direction for our future work.
\vspace{-10pt}
\section{Acknowledges}
\vspace{-5pt}
This work was supported by National Science and Technology Major Project (2024ZD01NL00101), Natural Science Foundation of China (62271013, 62031013), Guangdong Provincial Key Laboratory of Ultra High Definition Immersive Media Technology (2024B1212010006), Guangdong Province Pearl River Talent Program (2021QN020708), Guangdong Basic and Applied Basic Research Foundation (2024A1515010155), Shenzhen Science and Technology Program (JCYJ20240813160202004, JCYJ20230807120808017), Shenzhen Fundamental Research Program (GXWD20201231165807007-20200806163656003).
\vspace{-10pt}
{
    \small
    \sloppy  
    \bibliographystyle{ieeenat_fullname}
    \bibliography{main}
    \fussy   
}
\vspace{-5pt}
\end{document}